\documentclass[a4paper,12pt]{article}
\usepackage{amssymb}
\usepackage{amsfonts}
\usepackage{amsmath}
\usepackage{amsthm}
\usepackage{url}
\usepackage{graphicx}
\usepackage{natbib}

\bibpunct{(}{)}{;}{a}{,}{,}

\newcommand{\R}[1]{\ensuremath{\mathbb{R}^{#1}}}
\newcommand{\REM}[1]{}

\begin{document}
\title{Multi-Layer Perceptrons and Symbolic Data}
\author{Fabrice Rossi$^{\star\dagger}$ and Brieuc Conan-Guez$^\bullet$}
\date{}
\maketitle

\begin{flushleft}
$^\star$ Projet AxIS, INRIA Rocquencourt, \\
Domaine de Voluceau, Rocquencourt, B.P. 105,\\
78153 Le Chesnay Cedex -- France
\end{flushleft}
\begin{flushleft}
$^\dagger$ CEREMADE, UMR CNRS 7534, Universit\'e Paris-Dauphine,\\
Place du Mar\'echal De Lattre De Tassigny,\\
75775 Paris Cedex 16 -- France
\end{flushleft}
\begin{flushleft}
$^\bullet$ LITA, Universit\'e de Metz, \\
Ile du Saulcy, \\
57045 Metz Cedex 1 -- France
\end{flushleft}

\begin{abstract}
  In some real world situations, linear models are not sufficient to represent
  accurately complex relations between input variables and output variables of
  a studied system. Multilayer Perceptrons are one of the most successful
  non-linear regression tool but they are unfortunately restricted to inputs
  and outputs that belong to a normed vector space.  In this chapter, we
  propose a general recoding method that allows to use symbolic data both as
  inputs and outputs to Multilayer Perceptrons. The recoding is quite simple
  to implement and yet provides a flexible framework that allows to deal with
  almost all practical cases. The proposed method is illustrated on a real
  world data set. 
\end{abstract}

\section{Introduction}
Multilayer Perceptrons (MLPs) are a powerful non-linear regression tool
\citep{Bishop95}. They are used to model non linear relationship between
quantitative inputs and quantitative outputs.  Discrimination is considered as
a special case of regression in which the output predicted by the MLP
approximates the probability for the input to belong to a given class.

Unfortunately, MLPs are restricted to inputs and outputs that belong to a
normed vector space such as \R{n} or a functional space \citep[see][for
instance]{RossiConanGuez05NeuralNetworks,RossiEtAl05Neurocomputing}.  In this
chapter, we propose a solution that allows to use MLP for symbolic data both
as inputs and as outputs.

\section{Background}
We briefly recall in this section some basic definitions and facts about
Multilayer Perceptrons. We refer the reader to \citet{Bishop95} for a much more
detailed presentation of neural networks. 

\subsection{The Multilayer Perceptron (MLP)}
A MLP is a flexible and powerful statistical modeling tool based on the
combination of simple units called neurons. 

More precisely, a neuron with $n$ real valued inputs is a parametric
regression model with $n+1$ real parameters given by:
$E(Y|X)=N(X,\alpha)=T(\alpha_0+\sum_{j=1}^n\alpha_jX_j)$. In this equation,
$Y$ is the target variable in \R{}, $X$ is the explanatory variable in \R{n}
($X_j$ is the $j$-th coordinate of $X$), $\alpha$ is the parameter vector in
\R{n+1} and $T$ is a fixed nonlinear function called the activation function
of the neuron. In this very basic regression model, $\alpha$ is the only
unknown information.

A MLP is obtained by combining neurons into layers and then by connecting
layers. A layer simply consists in using several neurons in parallel, in
general with the same activation function for all neurons. We build this way a
multivariate regression model given by:
$E(Y|X)=H(X,\alpha_1,\ldots,\alpha_p)$. In this model, $Y$ is now a target
variable with values in \R{p}. Each coordinate of $Y$ is modeled by a simple
neuron, i.e.  $E(Y_i|X)=T(\alpha_{i,0}+\sum_{j=1}^n\alpha_{i,j}X_j)$.

We then combine layers by a simple composition rule. Assume for instance that
we want to build a multivariate regression model of $Y$ (with values in \R{p})
on $X$ (with values in \R{n}). A possible model is obtained with a two layer
MLP using $q$ neurons in its first layer and $p$ neurons in its second layer.
In this situation, the model is given by
$E(Y_i|X)=T_2(\beta_{i,0}+\sum_{k=1}^q\beta_{i,k}Z_k)$, where the $q$
intermediate variables $Z_1,\ldots,Z_q$ are themselves given by
$Z_k=T_1(\alpha_{k,0}+\sum_{j=1}^n\alpha_{k,j}X_j)$. In fact, the $Z_k$
variables are obtained as outputs from the first layer and used as inputs for
the second layer. Obviously, more than two layers can be used.

In this regression model, the activation functions $T_1$ and $T_2$ as well as
the number of neurons $q$ are known parameters, whereas the vectors $\alpha$
and $\beta$ have to be estimated using the available data. 

\subsection{Training and model selection}
Given a sample of size $N$, $(Y^i,X^i)_{1\leq i\leq N}$ distributed like
$(Y,X)$, our goal is to build a model that explains $Y$ thanks to $X$, i.e.,
we want to approximate $E(Y|X)$. On a theoretical point, this can be done with
a MLP \citep[see for instance][]{Whit}. 

Let us first consider a fixed architecture, that is a fixed number of layers,
with a fixed number of neurons in each layer and with a given activation
function for each layer. In this situation, we just have to estimate the
numerical parameters of the model. Let us call $w$ the vector of all numerical
parameters of the chosen MLP (parameters are also called weights). The
regression model is $E(Y|X)=H(X,w)$. As $E(Y|X)$ can be distributed
differently from $H(X,w)$, it is possible that for all $w$, $H(X,w)$ is
distinct from $E(Y|X)$, therefore, we have to choose $w$ so as to minimize
the differences between $H(X,w)$ and $E(Y|X)$. This is done indirectly by
choosing an error measure in the target space (\R{p}), denoted $d$ and by
searching for $w$ that minimizes $\mathcal{E}(w)=E(d(Y,H(X,w)))$.

In practice, this is done by minimizing the empirical error defined by:
\[
\widehat{\mathcal{E}}_N(w)=\frac{1}{N}\sum_{i=1}^Nd(Y^i,H(X^i,w))
\]
This minimization is performed by a gradient descent algorithm, for instance
the BFGS method or a conjugate gradient method \citep[see][]{Pres,Bishop95}. 

In practice, we use for $d$ the quadratic distance or the cross-entropy,
depending on the setting: the rational is to obtain a maximum likelihood
estimate in cases where there is actually a $w$ such that $E(Y|X)=H(X,w)$. See
section \ref{subsectionOutput} for practical examples of the choice of the
error distance in the case of symbolic data.

Unfortunately, even if $w$ is optimal and minimizes
$\widehat{\mathcal{E}}_N(w)$, the model might be limited because the
architecture of the MLP was badly chosen. We have therefore to perform a model
selection to choose the number of neurons, and possibly the number of layers
and the activation functions. Traditional model selection techniques can be
used to perform this task. We can for instance compare different models using
an estimation of $\mathcal{E}(w)$ constructed thanks to $k$-fold
cross-validation, bootstrap, etc. \citep[see][]{Bishop95}. 

\section{A numerical coding approach}
As stated in the introduction, MLPs are restricted to real valued inputs and
outputs. Some extensions allow to use functional input
\citep[see][]{RossiConanGuez05NeuralNetworks,RossiEtAl05Neurocomputing}. Older
works have also proposed to use interval valued inputs \citetext{see
  \citealp{Sima95,Simoff96,Beheshti98}; \citealp[see also][and section
  \ref{subsectionAlternative}]{RossiConanIFCS2002}} but there is currently no
simple way to deal with symbolic data. In this chapter we propose a numerical
coding approach that allows to use MLP on almost arbitrary symbolic data.

\subsection{Recoding one symbolic variable}\label{subsectionOneRecoding}
In this section, we present a numerical coding scheme for each type of
variables. 
Single valued variables are first considered (quantitative and categorical variables),
then we focus on symbolic variables: 
\begin{itemize}
\item {\bf Quantitative single valued variable}

Obviously, we don't have to do any recoding for a quantitative single valued
variable as this is a standard numerical variable.

\item {\bf Categorical single valued variable}\nopagebreak[4]

  Values of a categorical single valued variable are categories (also called
  modalities). Let $\left\{A_{1}, \ldots, A_{m}\right\}$ be the list of those
  categories. A categorical single valued variable is recoded thanks to the 
  traditional disjunctive coding, as summarized in the following table:
\begin{center}
\begin{tabular}{|l|l|l|}
\hline
$A_{1}$  & is recoded as  & $(1,0,0, \ldots,0)$ \\ \hline 
$A_{2}$  & is recoded as  & $(0,1,0, \ldots,0)$ \\ \hline 
\vdots &  \vdots        & \vdots           \\ \hline 
\vdots &  \vdots        & \vdots           \\ \hline 
$A_{m}$  & is recoded as  & $(0,0,\ldots,0,1)$ \\ \hline 
\end{tabular}
\end{center}
Therefore, we replace the considered categorical single valued variable by $m$
numerical variables. It should be noted that if it exists an order between modalities
(if we have $A_1<A_2\ldots<A_m$), the disjunctive coding scheme is not well adapted
to the problem nature. In such case, a standard numerical coding, where each modality
is replaced by its rank ($A_i \rightarrow i$), should be considered. We obtain this way
a numerical variable. 

\newpage
\item {\bf Interval variable}

  An interval variable is described by a pair of extreme values, that is
  $[a,b]$. We replace one interval variable by two quantitative single valued
  variables, respectively $\mu=\frac{a+b}{2}$ (the mean of the interval) and
  $\delta=b-a$, the length of the interval (we refer to this coding as the
  ``mean and length coding''). On a statistical point of view, it is better to
  use $\log\delta$ rather than $\delta$ directly, but some symbolic data
  include zero length intervals and it is therefore not always possible to use
  the logarithmic representation (see section \ref{subsectionOutput}). Another
  possibility is to recode $[a,b]$ as $a$ and $b$, considered as quantitative
  single valued variable (this is the ``bound based coding''). 

\item {\bf Categorical multi-valued variable}

  Categorical multi-valued variables are generalizations of categorical single
  valued variables for which the value of the variable is no more a category
  but a subset of the set of categories. We use exactly the same coding
  strategy as above ($m$ numerical variables), but we allow several 1 for a given variable, one in each
  column corresponding to a category in the subset. For instance,
  $\left\{A_{1},A_{3}\right\}$ is recoded as $(1,0,1,0, \ldots,0)$.

\item {\bf Modal variable}

Modal variables are generalizations of categorical multi-valued variables for
which each category has a weight. We use again the $m$ variables coding but
we use the weights of the modalities rather than 0 and 1's. 
\end{itemize}

\subsection{Recoding the inputs}
Input data are recoded as explained in the previous section, but additional
care is needed. It is indeed well known that MLP inputs must be centered and
scaled before being considered for training in order to avoid numerical
problems in the training phase (when we minimize
$\widehat{\mathcal{E}}_N(w)$). Moreover, this preprocessing must be compatible
with the initialization strategy used by the minimizing algorithm. Indeed, all
gradient descent algorithms are iterative: we start from a randomly chosen
solution candidate $w_0$ and we improve its quality iteratively. In general,
$w_0$ uses small initial values and it is important to be sure that the MLP
inputs will belong to the same approximate range of values. It is therefore
important to apply centering and scaling to the recoded inputs before the
training. 

Moreover, it is very common in practice to use regularization in order to
improve the quality of the modeling performed by the MLP \citep[and to avoid
over-fitting see][]{Bishop95}. This is done by minimizing
a new error function $\widehat{\mathcal{R}}_N(w)$ rather than
the standard error $\widehat{\mathcal{E}}_N(w)$. The rational of this new
error is to penalize complex models. This can be done for instance thanks to
the following error function:
\begin{equation}
\widehat{\mathcal{R}}_N(w)=\widehat{\mathcal{E}}_N(w)+\sum_{j=1}^t\lambda_jw_j^2.
\end{equation}
In this equation, $\lambda_j$ is a penalty factor for weight $j$ and is called
the \textbf{weight decay parameter} for weight $j$. Its practical effect is to
restrict the allowed variation for this weight: when $\lambda_j$ is small, the
actual value of $w_j$ has almost no effect on the penalty term included into
$\widehat{\mathcal{R}}_N(w)$ and therefore this weight can take arbitrary
values. On the contrary, when $\lambda_j$ is big, $w_j$ must remain small.

In general, we use only one value and $\lambda_j=\lambda$ for all $j$, but in
some situations, we use one penalty term per layer (the optimal value of the
weight decay parameters is determined by the model selection algorithm). In
our situation, the recoding scheme introduces some problems. Indeed, a
 categorical variable with a lot of categories is translated into a lot of
variables. This means that the corresponding numerical parameters will be
heavily constrained by the weight decays. Therefore, the recoding method
introduces arbitrary differences between variables when we consider them on a
regularization point of view. In order to avoid this problem, we normalize the
decay parameter.

Let us consider for instance a categorical single valued variable with 5
categories translated into 5 variables $X_1,\ldots, X_5$. For each neuron in
the first layer, we have $5$ corresponding weights, for instance
$w_{1,1},\ldots, w_{1,5}$ for the first neuron, $w_{2,1},\ldots, w_{2,5}$ for
the second neuron, etc. The corresponding penalty is
$\widehat{\mathcal{R}}_N(w)$ is normally $\lambda\sum_{k=1}^q\sum_{i=1}^5
w_{k,i}^2$, if there are $q$ neurons in the first layer and if we use only
weight decay for the whole layer. We propose to replace this penalty by
$\frac{1}{5}\lambda\sum_{k=1}^q\sum_{i=1}^5 w_{k,i}^2$, that is we divide the
weight decay parameter used for each variable corresponding to the recoding of
a categorical single valued variable by the category number of this variable.

We use the same approach for extension of the categorical type, i.e. the
categorical multi-valued and the modal types. We don't modify the weight decay
for interval variables as they really are comparable to two variables: for
instance, modifying one and not the other is meaningful, which is not the case
for categorical and modal variables.

\subsection{Recoding the outputs}\label{subsectionOutput}
Output data are recoded as explained in section \ref{subsectionOneRecoding},
but additional care is again needed. First of all, a noise model has to be
chosen in order to justify the use of an error measure $d$. More precisely,
while any error measure with derivatives can be used, it is very important to
model the way $Y$ behaves around $E(Y|X)$ so as to obtain sensible estimation
of $w$, the weight vector. Given a model of $Y$ (for instance Gaussian with
known variance and mean given by $E(Y|X)$, i.e., the output of the MLP), we
can choose an error measure that leads to a maximum likelihood estimate for
$w$: in the case of numerical output, using a quadratic error corresponds to
assuming that the noise is Gaussian, which is sensible.

Second of all, some constraints must be enforced on the outputs of the MLP so
as to obtain valid symbolic values: the length of an interval must be
positive, the sum of values obtained for a modal variable must be one,
etc. 

For non numerical variables, the situation is quite complex. Let us review the
different cases:
\begin{itemize}
\item {\bf Interval variable} 
  
  We have here both a noise problem and a consistency problem. Indeed, while
  the mean of an interval can be arbitrary, this is not the case of its length
  that must be positive. Therefore, the output neuron that model the length
  variable obtained by recoding an interval must use an activation function
  that produces only positive values.  Moreover, while it is sensible to
  assume that the noise is Gaussian for the mean of the target interval
  leading to a quadratic error measure for the corresponding output, this
  assumption is not valid for the length. 

  A simple solution can be applied to symbolic data in which there is no zero
  length interval. In this situation, rather than recoding $[a,b]$ into two
  variables $\mu=\frac{a+b}{2}$ and $\delta=b-a$, we replace $\delta$ by
  $l=\log \delta$. With this transformation, we can use a regular activation
  function and model the noise as Gaussian on $l$. Therefore, the error
  measure can be the quadratic distance. 

  Unfortunately, this recoding is not possible for degenerate intervals
  ($[a,a]$) which might be encountered. In this kind of situation, we have to
  use an adapted activation function and to choose a model for the variability
  of $\delta$. A possibility is to use a Gamma distribution (or one of its
  particular cases such as the exponential distribution or a Chi-square
  distribution). This implies the use of a specific error measure.

  The case of a bound based recoding is similar. If we recode $[a,b]$ into two
  variables $a$ and $b$, we have to ensure that $b\geq a$. There is no direct
  solution to this problem and we have to rely on specific activation
  functions. It is therefore simpler to use the mean and length based
  recoding for output variables. 

\item {\bf Categorical single valued variable}
  
  This case has been already studied by the neural community because it
  corresponds to a supervised classification problem. Indeed, when we want to
  classify inputs into classes $A_1$ to $A_m$, we build a prediction function
  that maps an input into a label chosen in the set $\{A_1,\ldots,A_m\}$. This
  can be considered similar to the construction of a regression for a
  categorical single valued target variable with values in
  $\{A_1,\ldots,A_m\}$.

  In order to train a MLP, we must be able to calculate the gradient of
  $\widehat{\mathcal{E}}_N(w)$ and therefore, the activation functions must be
  derivable. As a consequence, a MLP cannot directly output labels. Of course,
  we will use the disjunctive coding proposed in section
  \ref{subsectionOneRecoding}, but the MLP will seldom output exact 0 and
  1. Therefore, we will interpret outputs as probabilities. 
  
  More precisely, let us assume that the target variable $Y$ is categorical
  single valued, with values in $\{A_1,\ldots,A_m\}$. It is therefore
  translated into $m$ variables $Y_1,\ldots,Y_m$ with values in $\{0,1\}$ and
  such that $\sum_{i=1}^mY_i=1$. Then the last layer of the MLP must have $m$
  neurons. Let us call $T_1,\ldots,T_m$, the outputs of the last layer. Using
  a \emph{softmax} activation function \citep[described below and
  in][]{Bishop95}, we can insure that $T_i\in[0,1]$ for all $i$ and that
  $\sum_{i=1}^mT_i=1$. The natural interpretation for those outputs is
  probabilistic: $T_i$ approximates $P(Y=A_i|X)$.
  
  The model for the recoded variable is normally
  $E(Y_i|X)=T_i=T(\beta_{i,0}+\sum_{k=1}^q\beta_{i,k}Z_k)$, where
  $Z_1,\ldots,Z_q$ are outputs from the previous layer. In order to build the
  \emph{softmax} activation function, we introduce
  $U_i=\beta_{i,0}+\sum_{k=1}^q\beta_{i,k}Z_k$ and we define $T_i$ by:
\[
T_i=\frac{\exp(U_i)}{\sum_{j=1}^m\exp(U_j)}
\]
This activation function implies that $T_i\in[0,1]$ and that
$\sum_{i=1}^mT_i=1$. Using the probabilistic interpretation, it is easy to
build the likelihood of $(T_1,\ldots,T_m)$ given the observation
$(Y_1,\ldots,Y_m)$. It is obviously:
\[
\prod_{i=1}^mT_i^{Y_i}
\]
The maximum likelihood principle leads to the minimization of the following
quantity: 
\[
d(Y,T)=-\sum_{i=1}^mY_i\ln T_i
\]
The corresponding distance is the cross-entropy, which should therefore be used
for nominal output.

To summarize, when we have a categorical single valued output variable with
$m$ categories:
\begin{itemize}
\item we use the disjunctive coding to represent this variable as $m$
  numerical variables;
\item we use a \emph{softmax} activation function for the corresponding $m$
  output neurons;
\item we use the cross-entropy distance to compare produced values to desired
  outputs;
\item the actual output of the MLP can be either considered directly as a
  model variable, or transformed into a categorical single valued variable by
  using a probabilistic interpretation of the outputs to translate
  numerical values into the most likely label. 
\end{itemize}

\item {\bf Categorical multi-valued variable}

  The case of categorical multi-valued variable is a bit more complex because
  such a variable does not contain a lot of information about the underlying
  data it is summarizing. Indeed if we have for instance a value of
  $\left\{A_{1},A_{3}\right\}$, it does not mean that $A_1$ and $A_3$ are
  equally likely. Therefore, we use a basic probabilistic interpretation: we
  assume that categories are conditionally independent knowing $X$.

  That said, the practical implementation is very close to the one used for
  categorical single valued variables. Let us indeed consider a categorical
  multi-valued target variable $Y$, with values in $\{A_1,\ldots,A_m\}$. It is
  translated into $m$ variables $Y_1,\ldots,Y_m$ with values in $\{0,1\}$ (the
  constraint $\sum_{i=1}^mY_i=1$ is no more valid). As for a nominal variable,
  we call $T_1,\ldots,T_m$, the outputs of the last layer of the MLP. We use
  an activation function such that $T_i\in [0,1]$, for instance the logistic
  activation function:
\[
T(x)=\frac{1}{1+\exp(-x)}
\]
Then, $T_i$ is interpreted as the probability that category $A_i$ appears in
$Y$. Given that categories are assumed independent, the likelihood
  of $(T_1,\ldots,T_m)$ given the observation $(Y_1,\ldots,Y_m)$ is again: 
\[
\prod_{i=1}^mT_i^{Y_i}
\]
As for a categorical single valued variable, the maximum likelihood estimation
is obtained by using the cross-entropy error distance. 

To summarize, when we have a categorical multi-valued output variable with $m$
categories:
\begin{itemize}
\item we use the 0/1 coding to represent this variable as $m$
  numerical variables;
\item we use an activation function with values in $[0,1]$ for the
  corresponding $m$ output neurons;
\item we use the cross-entropy distance to compare produced values to desired
  outputs;
\item we use a probabilistic interpretation of the outputs: we consider that
  categories $A_i$ belongs to the categorical multi-valued output if and only
  if $T_i>0.5$. 
\end{itemize}

\item {\bf Modal variable}
  
  The case of modal variable can be handled almost exactly as the one of a
  categorical single valued variable. Let us consider indeed a modal variable
  with support $\mathcal{A}=\{A_1,\ldots,A_m\}$. It is described by a vector
  of \R{m}, $(p_1,\ldots,p_m)$ with the following additional constraints:
\begin{itemize}
\item $p_i\in[0,1]$ for all $i$;
\item $\sum_{i=1}^mp_i=1$. 
\end{itemize}
A modal variable must be interpreted as a probability distribution on
$\mathcal{A}$. It is recoded by the vector $(p_1,\ldots,p_m)$. Unfortunately,
we don't know exactly how the probability distribution has been built. As
symbolic descriptions are often summaries, we will assume here that $l$
micro-observations with values in $\mathcal{A}$ were used to build the
estimated probabilities $(p_1,\ldots,p_m)$. This implies that $lp_i$ out of
$l$ observations correspond to the category $A_i$. 

Exactly as for a categorical single valued variable, we use $m$ output neurons
with a \emph{softmax} activation function. We denote again $T_1,\ldots,T_m$
the corresponding outputs. With the proposed interpretation of the variable,
the likelihood of $T_1,\ldots,T_m$ given the observation $(p_1,\ldots,p_m)$
is (by construction $lp_i$ are integers):
\[
\frac{l!}{(lp_1)!\ldots (lp_m)!} T_1^{lp_1}T_2^{lp_2}\ldots T_m^{lp_m}
\]
The maximum likelihood principle leads to the minimization of the following
quantity: 
\[
d(Y,T)=-l\sum_{i=1}^mp_i\ln T_i
\]
Of course, $l$ might be removed from this cross-entropy like distance, but
only if each considered value of the modal variable $Y$ comes from $l$
micro-observations. When the number of micro-observations depends on the value
of the variable, we must keep this weighting in the error distance.
Unfortunately, this value is not always available. When the information is
missing, we can use the cross-entropy error distance without weight.

To summarize, when we have a modal output variable with $m$
categories:
\begin{itemize}
\item we use the probabilities associated to the categories to translate the
  variable into $m$ real valued variables;
\item we use a \emph{softmax} activation function for the corresponding $m$
  output neurons;
\item we use the cross-entropy distance to compare produced values to desired
  outputs;
\item when the information is available, we use the size of the
micro-observations set that has been used to produce the modal
description as a weight in the cross-entropy distance;
\item thanks to the \emph{softmax} activation function, the output of the $m$
  neurons are probabilities and can therefore be directly translated into a
  modal variable. 
\end{itemize}
\end{itemize}

\subsection{Alternative solutions}\label{subsectionAlternative}
Alternative solutions for interval valued inputs have been proposed in earlier
works \citep{Sima95,Simoff96,Beheshti98}. The basic idea of these works is to
use interval arithmetic, an extension of standard arithmetic to interval
values \citep[see][]{Moore66}. The main interest of interval arithmetic is to
allow to take into account uncertainty: rather than working on numerical
values, we work on intervals centered on the considered numerical values. 

Unfortunately, these approaches are not really suited to symbolic data. We
showed in \citep{RossiConanIFCS2002} that a recoding
approach provides better results than an interval arithmetic approach. The
main reason is that extreme values in an interval do not always correspond to
uncertainty. In meteorological analysis for instance, we cannot differentiate
broad classes of climate simply by using the mean temperature: extreme values
give valuable information, as a continental weather corresponds in general to
important yearly variation around the mean, whereas oceanic weather
corresponds to smaller yearly variation (see also section
\ref{sectionExperiments}).

\section{Open problems}
\subsection{High number of categories}
It is unfortunately common to deal with categorical variables (or modal
variables) with a lot of categories. The proposed recoding method introduces a
lot of variables. The practical consequence is a slow training phase for the
MLP. In some situations, when the number of categories is really high (100 for
instance), this might event prevent the training from succeeding.

A possibility is to use a lousy encoding in which several categories are
merged, for instance based on their frequencies in the data set.
Unfortunately, it is very difficult to do this kind of simplification while
taking into account target variables. Indeed, an optimal unsupervised lousy
encoding might loose small details that are needed for a good prediction of
target variables.

\subsection{Multiple outputs}
We have shown in section \ref{subsectionOutput} how to deal with symbolic
outputs. While we can handle almost all type of symbolic data, we have to be
extremely careful when mixing symbolic outputs of different types. For
instance, it is well known \citep[see][]{Bishop95} that using the quadratic
error distance for vector output corresponds to assuming that the noise is
Gaussian, with a fixed variance and independent on each output. Departure from
this model (for instance a different variance for each output) is possible but
implies the modification of the error measure. 

The problem is even more crucial when we mix different types of symbolic data.
If we have for instance a numerical variable and a categorical single valued
variable, simply using the sum of a quadratic distance and of a cross-entropy
distance will seldom result in a maximum likelihood estimation of the
parameters of the MLP. One has at least to take into account the variance of
the noise of the numerical variable. Moreover, the basic solution will be to
assume that the outputs are conditionally independent given the input, but
this might be a very naive approach.

We do not believe that there is an automatic general solution for this
problem and we insist on the importance of choosing a probabilistic model for
the outputs in order to obtain meaningful results. 

\subsection{Other types of symbolic data}
Our solution does not deal with additional structure in symbolic data. For
instance, we don't take into account taxonomies or rules. A way to deal with
taxonomies is to use a hierarchical recoding: the deepest level of the
hierarchy is considered as a set of category for a categorical single valued
variable which leads to a disjunctive coding. Values from higher levels of the
hierarchy are coded as categorical multi-valued values that is by setting to 1
all categories that are descendant of the considered value.

Another limitation comes from the fact that we cannot deal with missing data:
if a symbolic description is not complete for one individual (for instance one
interval variable is missing for this individual), treatment of such
individual by the MLP model is not possible.  One solution to overcome this
limitation is to apply classical imputation methods on recoded variables,
i.e., to replace missing data by ``guessed'' values.  Of course, some care
have to be taken to respect the semantic of imputed variables.  A naive
method consists in replacing missing values by means of corresponding
variables.  A more sophisticated method relies on the k nearest neighbor
(k-NN) algorithm: given a vector in which some coordinates are missing, we
calculate its k nearest neighbors among vectors that do not miss these
coordinates, and we replace missing values by averages of coordinates of the k
nearest neighbors. Usually meta-parameter k is determined by
cross-validation. It is also possible to use this kind of imputation methods
directly at the symbolic level, i.e. before the recoding phase, if some
generalized mean operator is available for the considered data \citep[see][for
examples of such operators]{BockDiday2000}

\section{Experiments}\label{sectionExperiments}

\subsection{Introduction}
In this section, we show on a semi-synthetic example how in practice neural
net models and more precisely MLPs can process symbolic objects. Only the
specific case of interval variables will be considered in these experiments
(categorical multi-valued variables and modal variables will not be addressed 
here and require additional experiments). Results obtained in this specific
example will allow us to tackle three important issues relative to treatment
of data thanks to symbolic objects:

\begin{itemize}
\item {\em benefits of symbolic objects over standard approaches:} symbolic
  approaches allow to represent data thanks to richer and more complex
  descriptions compared to standard approaches (for instance the mean approach
  where data are simply averaged).  We have seen in the beginning of this
  chapter that these complex descriptions imply some specific adaptations of
  neural net models (adaptation of the activation function, careful use of the
  weight decay technique, etc.). Moreover, in some cases, the model complexity
  (which is related to the number of weights) can be higher in the case of
  symbolic approaches than in the case of standard approaches: indeed a
  categorical single valued variable with a high number of categories implies
  a high dimensional input, and therefore a high number of weights.
  As a direct consequence, estimation of neural net models can be more
  difficult when dealing with symbolic objects. It is therefore legitimate to
  wonder if symbolic approaches are worthwhile. In the proposed example, we
  show that they are indeed very helpful: model performances are 
  improved thanks to symbolic objects.

\item {\em difficult choice of the coding method:} in many real world
  problems, the practitioner has at its disposal the raw data, that is, primary
  data on which no preprocessing stages have been applied.  He is therefore
  free to choose the best coding method to recode such data into symbolic
  objects: for instance he can recode the raw data into modal variables, or
  into intervals according to the problem nature. For each of these recoding
  methods, some meta-parameters have to be set up: for instance in the case of
  modal variables, it is up to the practitioner to choose the number of
  categories.  In the case of interval variable, he has to decide which of the
  bound based coding and of the mean and length coding is more
  appropriate. As we will see in the proposed experiments, such choices have
  noticeable impact on model performances.

\item {\em low quality data and robustness of models:} finally, it is not
  uncommon in many real world problems to have to deal with low quality raw
  data, that is, data with missing values or with noisy measurements.  It is
  therefore tempting to wonder if symbolic approaches can cope successfully
  with such data. Once again, proposed experiments will allow to address this
  problem: symbolic approaches perform better when dealing with low quality
  data than standard approaches.

\end{itemize}

\subsection{Data Presentation}

In order to address the different issues described above, we have chosen a
synthetic example based on real world data: we consider climatic data from
China published by the Institute of Atmospheric Physic of Chinese Academy of
Sciences (Beijing) \citep[see][]{ChineseWeatherDataSet1997}.  This application
concerns mean monthly temperatures, and total monthly precipitations observed
in 260 meteorological stations spread over the Chinese territory. In these
experiments, we restrict ourselves to the year 1988: each station is therefore
described by a single vector (12 temperatures and 12 precipitations).
Moreover, we have the coordinate of all the stations (longitude and latitude).

As the goal of this chapter is to show how in practice we can process symbolic
objects thanks to neural net models, we represent meteorological information
related to each station thanks to symbolic objects (only interval variables
are considered in this application).  The goal of these experiments is then to
infer from the meteorological description of each station (which can be
symbolic or not), its location in China (longitude and latitude). This problem
is not a real practical application, but it has two interesting
characteristics:
\begin{itemize}
\item first, inference of station location is quite a difficult task, as it is
  obviously a ill-posed problem. Indeed, we try to model thanks to a MLP the
  inverse of the function which maps station location to meteorological
  description.  As this function is not a one to one mapping, (two stations
  located far away one from the other in China can have very similar
  meteorological descriptions), the inverse function is therefore a set-valued
  function, which is very difficult to model thanks to standard techniques.
  We will see that MLPs based on symbolic objects perform better in this case
  than standard approaches.

\item secondly, the data are not native symbolic data and we have access to
  underlying raw data. This allows to study the effect of the coding on model
  performances.  Moreover, as we are interested in robustness of symbolic
  approaches when dealing with low quality data, we can intentionally degrade
  the original data, and then study consequences on model performances (we
  remove from original data some chosen values).  Experiments will show that
  thanks to symbolic approaches, models estimated on high quality data (data
  with no missing values), have correct performances on low quality data.
\end{itemize} 

\subsection{Recoding Methods and Experimental Protocol}

We consider four different experiments corresponding to standard and symbolic
approaches (the first two one can be considered as standard approaches,
whereas the last two one are symbolic approaches):
\begin{itemize}
\item in the first experiment, we simply process the raw data. Therefore, the
  input of the MLP model is a vector of 24 coordinates (the 12 temperatures and
  the 12 precipitations associated to a given station over one year).  It is
  worth noticing that in this experiment the model complexity is quite high,
  as the number of weights is directly linked to the input
  dimension (24 in this case).
\item in the second experiment, we aggregate temperature data and
  precipitation data thanks a simple averaging. Therefore, in this case, each
  station is described by a two-dimensional vector:
  $(temp_{mean},precip_{mean})$.
\item in the third experiment, temperature data as well as precipitation data
  are recoded into intervals. In this case, each station is described by its
  extrema: $([temp_{min},temp_{max}],[precip_{min},precip_{max}])$.  Input
  dimension of the MLP model is 4, as each pair of intervals is submitted as a
  four dimensional vector.
\item finally, in order to explore different coding methods, we submit to the
  MLP model the mean and the standard deviation of each variable. This
  corresponds to a robust interval coding in which the length is estimated
  thanks to the standard deviation rather than using the actual extreme
  values. Each station is therefore described by
  $(temp_{mean},temp_{sd},precip_{mean},precip_{sd})$.  Input dimension of
  each MLP is 4.
\end{itemize} 

In all the experiments, we use two distinct MLPs: one for the longitude
inference, and one for the latitude inference.  Of course, it would have been
possible to infer the location (longitude and latitude) as a whole by a unique
MLP. As we will see in the experiments, such approach is not well adapted to
the problem nature: indeed, latitude inference is much easier than longitude
inference. Therefore, in order not to penalize one problem over the other, we
decided to keep problems separated.

All the MLPs in this application have a single hidden layer.  We test
different size for the hidden layer: 3, 5, 7, 10, 15, 20, 30 and 40 hidden
neurons.  In order to estimate models and to compute a good estimate of their
real performances, the whole data set is split into three parts: the training
set which contains 140 stations.  This set is used to estimate model
parameters thanks to a standard minimization algorithm (a conjugate gradient
method).  The error criterion is the quadratic error.  The second part of the
data set is the validation set (60 stations) which is used to avoid
over-fitting: the minimization algorithm is stopped when the quadratic error on
this validation set is minimum.  For each experiment, the minimization is
carried out 10 times: the starting point is randomly chosen at each time. The
best architecture (the number of hidden neurons and the values of the weights)
is chosen according to the quadratic error on the validation set.  Finally,
as error on validation set is not a good estimate of model real performances,
we compute the mean absolute error in degree on the test set (60 stations).

Table \ref{errorDegree} summarizes performances obtained in the different
experiments:

\begin{table}[htbp]
  \centering
\begin{tabular}{|c|c|c|c|}\hline
Inputs         & Longitude & Latitude  & Number of weights \\\hline
Full data (24) & 4.07 (3)  & 1.27 (30) & 860                        \\\hline
Mean           & 7.31 (30) & 2.51 (17) & 190                        \\\hline
Mean \& StdDev & 4.91 (20) & 1.34 (25) & 272                        \\\hline
Min \& Max     & 4.73 (25) & 1.56 (40) & 392                        \\\hline
\end{tabular}
 \caption{Mean absolute error in degree and architecture complexity}
  \label{errorDegree}
\end{table}

These results clearly show that longitude inference is a more difficult task
than latitude one: indeed, in the latter, model accuracy is close to 1
degree, whereas in the former the accuracy is at best 4 degrees. On the whole
data set , the range for longitude is 56 degrees and the range for latitude is
34.23 degrees. This means that the best relative error for latitude is less
than 4\% whereas it is more than 7\% for longitude.

This difference can be partly explained by analyzing precisely characteristics
of the Chinese climate: due to its large surface, it exists a large diversity
of climates in China (cold dry weather to hot wet weather).  If we focus on
the temperature and the precipitation, some geographical inference can be
made.  The mean yearly temperature is of course very informative on the
station latitude (south is hot, north is cooler). However, the coldest region
of China is Tibet (south-west), which is not located in the north of China.
The total yearly precipitation is informative on the axis Xinjiang
(north-west) - Guangzhou (south-east): The Xinjiang region is very dry,
whereas Guangzhou city has a very wet climate (monsoon).  Therefore we can see
that both variables contribute quite obviously to the inference of the
latitude.  For the longitude case, accuracy is not as good, as both variables
don't bring as much information on this axis east-west.

If we study now performances of the different approaches, we can see that the
full data approach gives the best performances. Performances of symbolic
approaches ($(min,max)$ and $(mean,sd)$) are quite good too, and very close to
those of the full data approach.  Finally the mean approach gives the worst
results. This leads to the following remarks:
\begin{itemize}
\item first, only approaches which are able to model the variability of the
  meteorological phenomenon over one year can achieve good performances.
  Indeed, if two distant stations have similar mean temperature and mean
  precipitation, then the mean approach can't infer their locations
  accurately.  This is the case for example for the cities of Urumqi and
  Harbin.  Urumqi is located in the north-west of China, and Harbin is located
  in the north-east of China.  Urumqi has a continental climate with very hot
  summer and very cold winter, whereas Harbin has an oceanic climate with cool
  summer and chilly winter.  However, both cities have quite similar mean
  temperature and mean precipitation over one year.  Therefore, the mean
  approach can't infer accurately their location.  In the case of the other
  approaches (full data approach and symbolic approaches), the variability of
  the meteorological phenomenon is preserved in the description (for instance,
  as explained before, for the temperature description, interval length is
  greater for continental climates than for oceanic climates), which leads to
  better performances.

\item Even if the full data approach leads to some slight performance
  improvements over symbolic approaches, the latter should be preferred over
  the former in practical situations.  Indeed, in the case of the full data
  approach, input dimension is quite high (input dimension is 24), which
  implies a high number of parameters for the model (860 weights).
  In the case of symbolic approaches, important information available in
  original data, such as variability, have been summarized thanks to a
  compact description. We can see that this data reduction doesn't impaired
  too much performances.  Moreover, model complexity is lower compared to the
  full data approach (272 for the $(mean,sd)$ coding and 392 for the
  $(min,max)$ coding), which leads to faster estimation phase.

\item finally, no advantages appears between the two symbolic representation.
  The $(min,max)$ coding and $(mean,sd)$ coding give quite similar results. We
  will see nevertheless in the next section, that both coding are not strictly
  equivalent.
\end{itemize}
    
\begin{figure}[htbp]
  \centering
  \includegraphics[width=0.8\textwidth]{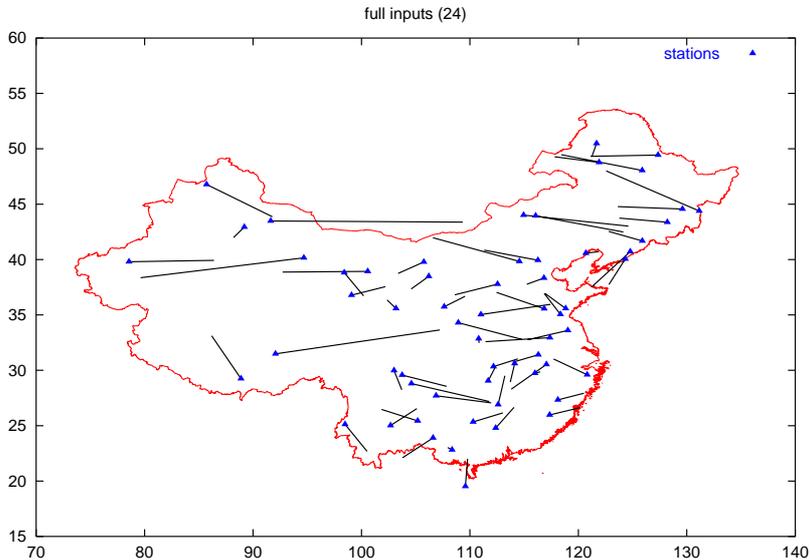}
  \caption{Model inferences vs true station locations: full data}
  \label{figFullData}
\end{figure}
    
\begin{figure}[hbtp]
  \centering
  \includegraphics[width=0.8\textwidth]{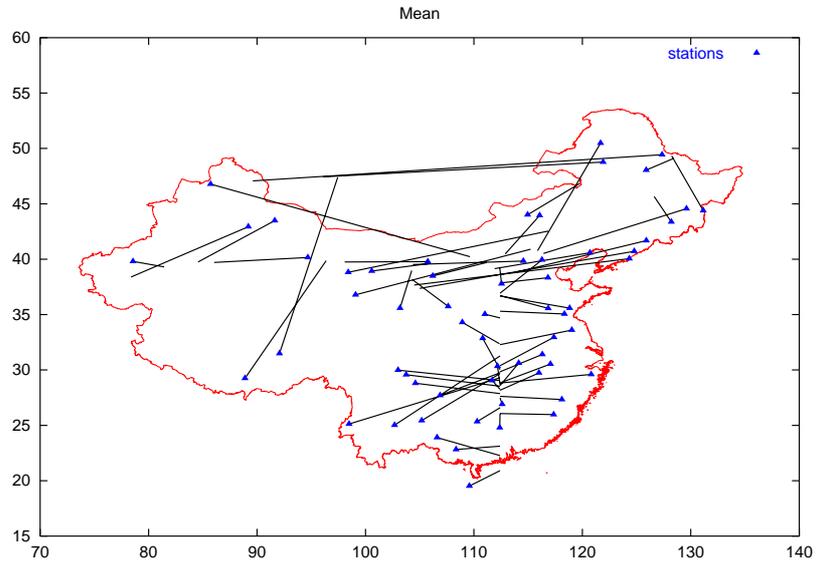}
  \caption{Model inferences vs true station locations: mean approach}
  \label{figMean}
\end{figure}

\begin{figure}[htbp]
  \centering
  \includegraphics[width=0.8\textwidth]{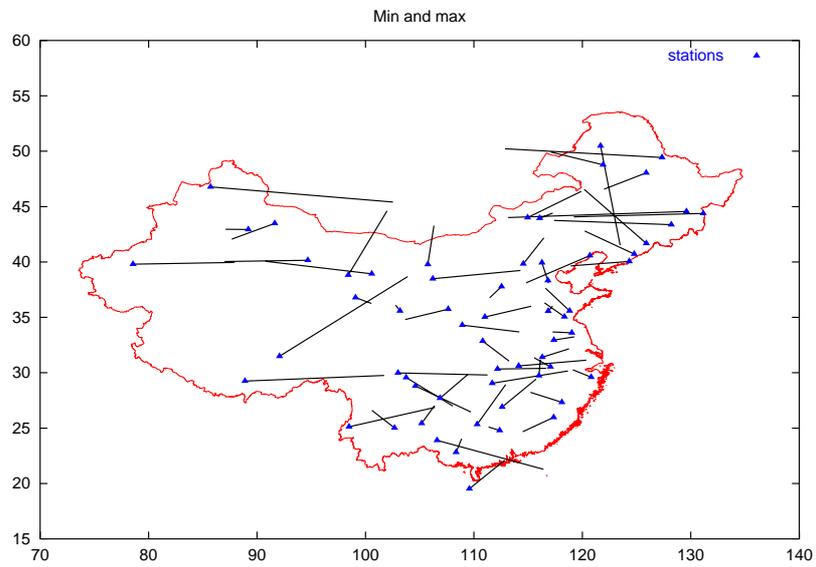}
  \caption{Model inferences vs true station locations: min \& max}
  \label{figMinMax}
\end{figure}

\begin{figure}[htbp]
  \centering
  \includegraphics[width=0.8\textwidth]{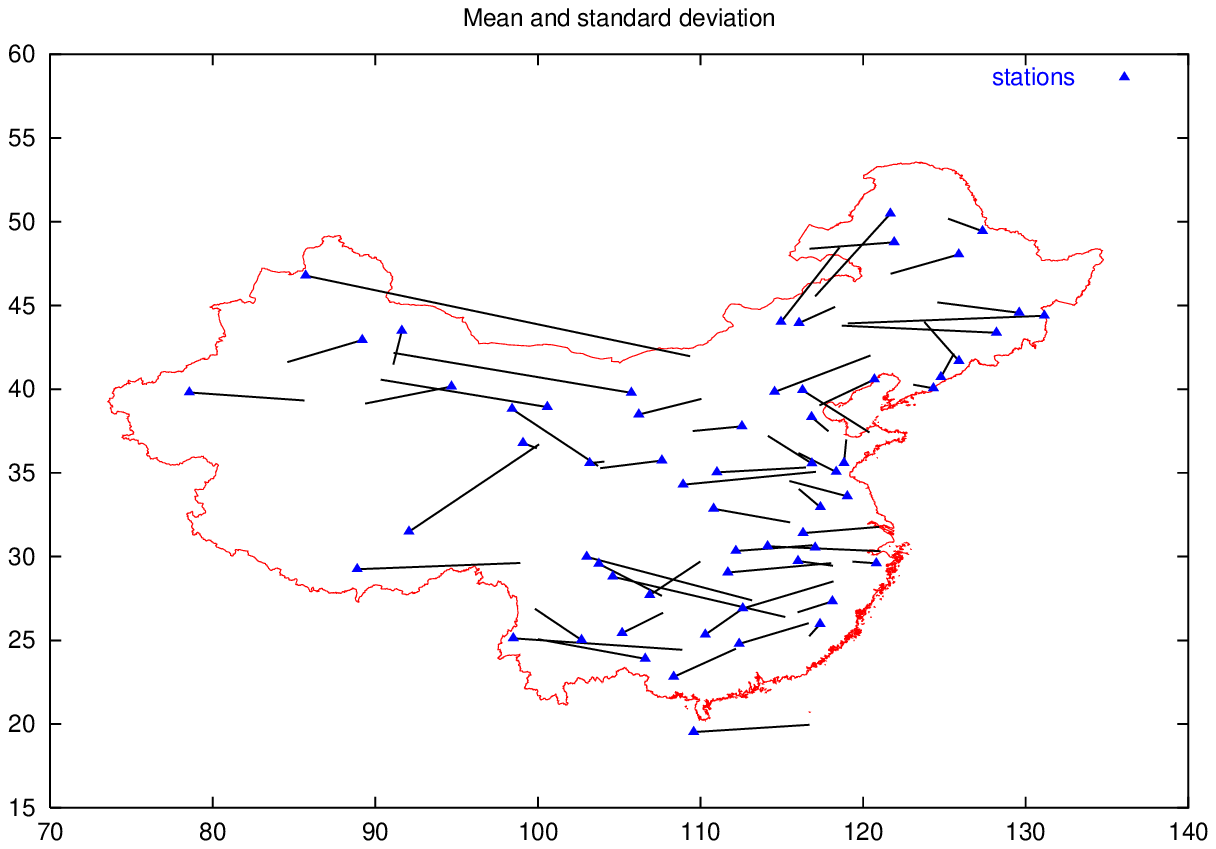}
  \caption{Model inferences vs true station locations: mean \& standard deviation}
  \label{figMom2}
\end{figure}

In order to illustrate the behavior of the different approaches, we represent
for each approach model inferences versus true station locations on different
maps (figures \ref{figFullData},\ref{figMean},\ref{figMinMax},\ref{figMom2}).
Triangles represent the location of the 60 stations which belong to the test
set. Segments represent the location inferred by the model.  In each figure, we
can see that segments tend to be horizontal, which corroborates the fact
that longitude inference is more difficult than latitude one.
 
\subsection{Low Quality Data} 
  
We have seen in the previous section, that the full data approach and symbolic
approaches achieve quite similar performances on the Chinese data example.
The only advantage at this point of symbolic approaches is that the model
complexity (number of weights) is lower in this case. The goal of
this section is to show that models based on symbolic objects are much more
robust to low quality data (data with missing values or with noisy
measurements) than those based on the full data approach.  More precisely, for
each approach (standard and symbolic), we consider the MLP model estimated in
the previous section: this estimation was carried out with high quality data
(no missing values).  Our goal is to study performances of this model when
unseen data with missing values (low quality data) are submitted.

In order to investigate the robustness of the different approaches, we
intentionally degrade the Chinese data which belong to the test set (the
training set and the validation set are not considered in this section, as we
use the models which have been estimated in the previous section).  Data
degradation is done 3 times: the first degradation leads to data with
mid-quality (half values are missing). The second degradation leads to low
quality data (two thirds are missing), and finally the last degradation leads
to very low quality data (three quarters are missing).  Data degradation is
done according to the following protocol: we consider for each meteorological
station the temperature vector (12 coordinates) and the precipitation vector
(12 coordinates).  For the first degradation, we remove regularly one coordinate
out of 2 for each vector (coordinates 2, 4, 6, 8, 10, 12 are missing values).
Therefore the temperature vector is now a 6-dimensional vector, just as the
precipitation vector.  For the second degradation, we remove regularly 2
coordinates out of 3 from the original data, and the dimension of each vector
is 4 (coordinates 2, 3, 5, 6, 8, 9, 11, 12 are missing values).  Finally, in
the third experiment we remove regularly 3 coordinates out of 4, and the
dimension of each vector is 3 (coordinates 2, 3, 4, 6, 7, 8, 10, 11, 12 are
missing values).

Models based on mean, min-max or mean-standard calculation can be applied
directly to these new data. All we have to do is to recompute each of these
quantities with the remaining values. For the full data approach, direct
treatment is not so simple, as MLP models have an input dimension of 24, which
is incompatible with the data dimension (12 for the first degradation, 8 for
the second degradation and 6 for the third degradation).  Therefore, in order
to submit these data to the full data model, we must replace each missing
value by an estimate.  Many well-known missing value techniques can be applied
in order to compute these estimates.  In order not to penalize the full data
approach, we choose to replace missing values by new values computed thanks to
linear interpolation.  For sake of clarity, we denote $c_i$ the $i^{th}$
coordinate.  For the first degradation, coordinates $c_2$, $c_4$, $c_6$, $c_8$,
$c_{10}$, $c_{12}$ are missing.  Coordinate $c_2$ is replaced by $(c_1+c_3)/2$.
We proceed the same way for coordinates $c_4$, $c_6$, $c_8$, $c_{10}$.  For
coordinate $c_{12}$, we make use of the periodic aspect of the climate:
coordinate $c_{12}$ is replaced by $(c_1+c_{11})/2$ (December is computed
thanks to November and January of the same year).  For the second degradation,
we have $c_2=(2c_1+c_4)/3$ and $c_3=(c_1+2c_4)/3$, and so on for coordinates
$c_5$, $c_6$, $c_8$, $c_9$, $c_{11}$, $c_{12}$.  Finally for the third
degradation, we have $c_2=(3c_1+c_5)/4$, $c_3=(c_1+c_5)/2$ and
$c_4=(c_1+3c_5)/4$, and so on for coordinates $c_6$, $c_7$, $c_8$, $c_{10}$,
$c_{11}$, $c_{12}$.

\begin{figure}[htb]
  \centering
  \includegraphics[width=13cm]{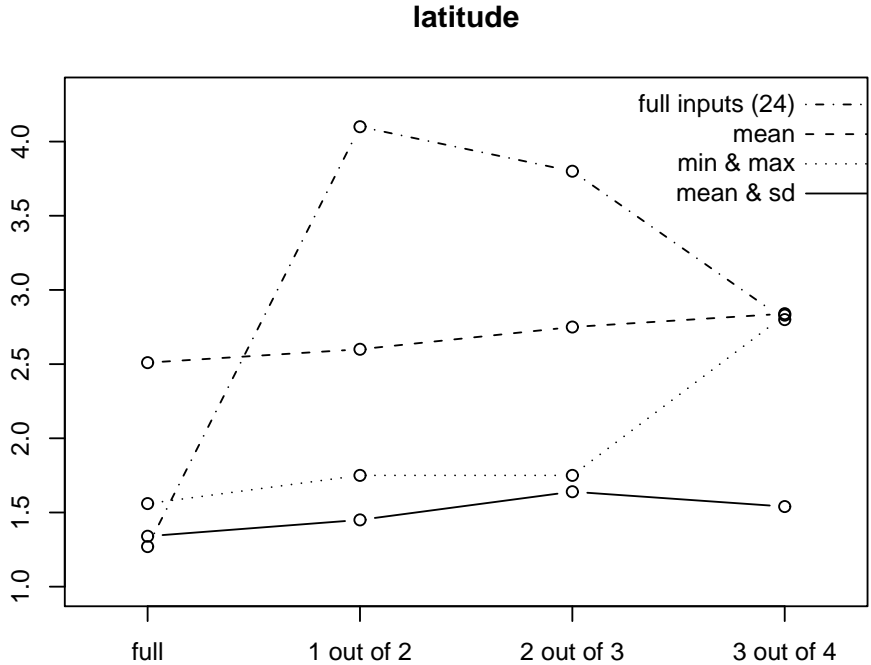}
  \caption{data quality reduction: latitude}
  \label{figLatitude}
\end{figure}
\begin{figure}[htb]
  \centering
  \includegraphics[width=13cm]{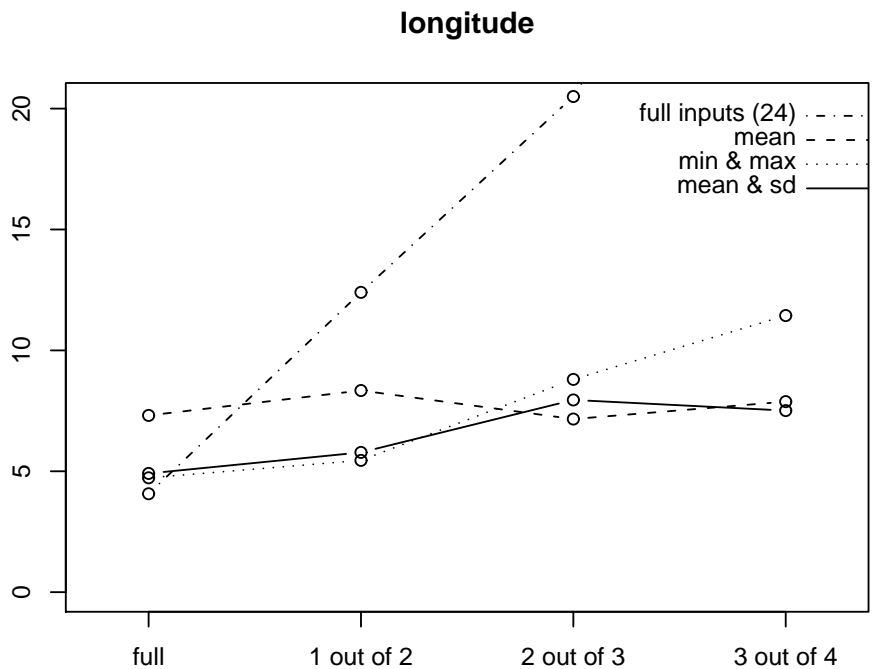}
  \caption{data quality reduction: longitude}
  \label{figLongitude}
\end{figure}

    
\noindent For each of the models estimated in the first experiments, we
compute the mean absolute error on the modified test set. Figure
\ref{figLatitude} summarizes performances obtained by the different approaches
for the inference of station latitude in respect to the data degradation
level.  Figure \ref{figLongitude} do the same for the longitude.
    
Results obtained by the different approaches leads to the following remarks:
\begin{itemize}
\item first we can see that the full data approach is very sensitive to data
  quality compared to the other approaches. Indeed, data degradation impairs
  strongly performances (error evolves from 4.07 degrees to 12.4 degrees for
  the longitude when half values are missing, and meanwhile error evolves from
  1.27 degrees to 4.1 degrees for the latitude).  Other approaches are not
  impacted in these proportions for the same degradation.  We can see there
  the outcome of using complex models, that is, models with a large number of
  weights: if the submitted data are close to the training data, the model
  achieve good performances. However, if submitted data differs too much from
  the training data, performances fall down.
\item we can see that performances achieved by the mean approach are quite
  uniform in respect to data degradation. Nevertheless, symbolic approaches
  outperform the mean approach in almost all cases. Once again, descriptions
  obtained thanks to a simple averaging are too poor, which avoid MLP models
  from making accurate inferences.
\item finally, we can see that the symbolic approach based on the $(mean,sd)$
  estimation is more robust than the one based on the $(min,max)$ estimation.
  We can explain this result by the fact that the $(min,max)$ estimation is
  more sensitive to out-layers (month with unusual temperature for instance)
  than the $(mean,sd)$ estimation.  This dependency has noticeable
  consequences on model inferences. We can conclude therefore that the
  $(mean,sd)$ estimation should be preferred in all cases.
\end{itemize}

\section{Conclusion}
We have proposed in this chapter a simple recoding solution that allows to use
arbitrary symbolic inputs and outputs for multilayer perceptrons. We have
shown that traditional techniques, such as weight decay regularization, can be
easily transposed to the symbolic framework.  Moreover the proposed approach
doesn't necessitate specific implementation: standard neural net toolbox can
be used to process symbolic data.  Experiments on semi-synthetic data have
shown that neural processing of intervals gives satisfactory results.  In
future works, we plan to extend these experiments to categorical multi-valued
variables and modal variables in order to validate all the recoding solutions.
Finally, we plan also to study more precisely the adaptation of standard
imputation methods (mean approach and k nearest neighbor approach) to the
symbolic framework, especially by comparing imputation on recoded variables
versus imputation on symbolic variables using symbolic mean operators. 

\bibliographystyle{abbrvnat}
\bibliography{smlp-chapter}

\end{document}